%% file: root.tex
\documentclass[a4paper,conference]{IEEEtran}
\IEEEoverridecommandlockouts
\usepackage{cite}
\usepackage{amsmath,amssymb,amsfonts}
\usepackage{algorithmic}
\usepackage{graphicx}
\usepackage{textcomp}
\usepackage{xcolor}
\usepackage{pxfonts} 
\usepackage{mathtools}
\usepackage{tikz}
\usetikzlibrary{arrows}
\usepackage{float}
\usepackage{url}
\usepackage{hyperref}

\input{latex-math/basic-ml}

\input{notation}

\newtheorem{proposition}{Proposition}
\newtheorem{theorem}{Theorem}

\newtheorem{definition}{Definition} 
\newtheorem{lemma}{Lemma}

\newcommand{\eqtext}[1]{\mathrel{\overset{\makebox[0pt]{\mbox{\normalfont\tiny\sffamily #1}}}{=}}}

\usepackage{ulem}
\usepackage{color}

\def\BibTeX{{\rm B\kern-.05em{\sc i\kern-.025em b}\kern-.08em
    T\kern-.1667em\lower.7ex\hbox{E}\kern-.125emX}}

\begin{document}

\title{Relative Feature Importance\\
\thanks{This work is funded by the German Federal Ministry of Education and Research (BMBF) under Grant No. 01IS18036A and supported by the Bavarian State Ministry of  Science  and  the  Arts  in  the  framework  of  the  Centre  Digitisation.Bavaria (ZD.B). The authors of this work take full responsibility for its content.}
}

\author{\IEEEauthorblockN{Gunnar K\"onig$^{1,2}$, Christoph Molnar$^1$, Bernd Bischl$^1$, Moritz Grosse-Wentrup$^{2,3,4}$}
\IEEEauthorblockA{
$^1$Institute for Statistics, LMU Munich, $^2$Research Group Neuroinformatics, University of Vienna,\\ $^3$Research Platform Data Science @ Uni Vienna, $^4$Vienna Cognitive Science Hub\\}}

\maketitle

\begin{abstract}
Interpretable Machine Learning (IML) methods are used to gain insight into the relevance of a feature of interest for the performance of a model. Commonly used IML methods differ in whether they consider features of interest in isolation, e.g., Permutation Feature Importance (PFI), or in relation to all remaining feature variables, e.g., Conditional Feature Importance (CFI). As such, the perturbation mechanisms inherent to PFI and CFI represent extreme reference points. We introduce Relative Feature Importance (RFI), a generalization of PFI and CFI that allows for a more nuanced feature importance computation beyond the PFI versus CFI dichotomy. With RFI, the importance of a feature relative to any other subset of features can be assessed, including variables that were not available at training time. We derive general interpretation rules for RFI based on a detailed theoretical analysis of the implications of relative feature relevance, and demonstrate the method's usefulness on simulated examples.
 \end{abstract}

\begin{IEEEkeywords}
feature importance, interpretable machine learning, explainable artificial intelligence, causality
\end{IEEEkeywords}

\section{Introduction}

Predictive modelling is increasingly deployed in high-stakes environments, e.g., in the criminal justice system \cite{Dressel2018}, loan approval \cite{Xia2017}, recruiting \cite{Dastin2018} and medicine \cite{Topol2019}. Due to legal regulations \cite{Voigt2017,Torre2018} and ethical considerations, ML methods need not only perform robustly in such environments but also be able to justify their recommendations in a human-intelligible fashion. This development has given rise to the field of interpretable machine learning (IML) that involves studying methods that provide insight into the relevance of features for model performance, referred to as feature importance.\\
Prominent feature importance techniques include permutation feature importance (PFI) \cite{Breiman2001rf,Fisher2019} and conditional feature importance (CFI) \cite{Strobl2008,Fisher2019,Molnar2020}. PFI is based on replacing the feature of interest $X_j$ with a perturbed version sampled from the marginal distribution $P(X_j)$ while CFI perturbs $X_j$ such that the conditional distribution with respect to the set $R$ of remaining features $P(X_j|X_R)$ is preserved. The sampling strategy defines the method's reference point and therefore affects the method's implicit notion of relevance. While PFI quantifies the overall reliance of the model on the feature of interest, CFI quantifies its unique contribution given \textit{all} remaining features.\\
While both PFI and CFI are useful, they fail to answer more nuanced questions of feature importance. For instance, a stakeholder may be interested in the importance of a feature relative to a subset of features. Also, the user may want to know how important a feature is relative to variables that had not been available at training time.\\
We suggest relative feature importance (RFI) as a generalization of PFI and CFI that moves beyond the dichotomy between PFI, which breaks all dependencies with features, and CFI, which preserves all dependencies with features. In contrast to PFI and CFI, RFI is based on a perturbation that is restricted to preserve the relationships with a set of variables $G$ \textit{that can be chosen arbitrarily}. We show that RFI is (1) semantically meaningful and (2) practically useful.\\
We demonstrate the semantical meaning of RFI in Section \ref{sec:interpreting-rfi}. In particular, we derive general interpretation rules that link nonzero RFI to (1) the conditional dependence of the feature of interest with the target and non-conditioned features $\XRu$ given the conditioned variables $X_G$ in the data and (2) the conditional dependence of the input to the feature of interest $X_j$ with the model's prediction $\hat{Y}$ given fixed inputs to the remaining features $X_R$ (Theorem~\ref{theorem:nonzero-rfi-implications}). Furthermore, we show that a nonzero difference between $\rfijg$ and $\rfijgn$, with $N$ being an arbitrary set disjunct with $G$, implies the conditional dependence $X_j \dpd X_N | X_G$ (Theorem~\ref{theorem:rfi-diff}).\\
In Section \ref{sec:estimation}, we provide an implementation of RFI estimation that is based on recent results from the related knockoff research field \cite{Candes2018,Romano2019}. Furthermore, we translate the testing framework developed for conditional feature importance \cite{Watson2019} to RFI. We support our theoretical analysis and findings by various simulation studies in Section \ref{sec:examples}. In particular, we show that RFI can expose the indirect contribution of variables that are not directly used by the model but provide information via dependent variables  (Section \ref{sec:examples:indirect-influence}). Similarly, we show how RFI can be used to assess feature importance with respect to variables not included at training time (Section \ref{sec:examples:confounder}).

\subsection{Contributions and Related Work}

While conditioning on subsets of variables has been suggested before \cite{Strobl2008,Fisher2019}, the implications of this generalized variant of CFI have not yet been rigorously analyzed. Some IML methods perturb or hide subsets of features, e.g., in the context of multiple regression relative importance analysis is a model-specific technique that averages over all importances of models trained on feature subsets \cite{Budescu1993,Lipovetsky2001}. Model-agnostic, local approximations to the respective feature effect that avoid retraining and instead perturb subsets of features have also been proposed \cite{vstrumbelj2014explaining,Lundberg2017}. A very recent global, model-agnostic feature importance proposal called SAGE quantifies feature importance by perturbing multiple features \cite{Covert2020}.\\
While the aforementioned approaches are all based on removing several features to provide more nuanced insight into the model, our proposal only modifies the feature of interest. Our approach is model-agnostic and global, while most aforementioned approaches are model-specific or local. The exception is the global, model-agnostic SAGE \cite{Covert2020}, however the approaches are not only computationally but also semantically different. E.g. our method assigns an importance of zero for features that are not used by the model\footnote{A proof of this property is given in Lemma~\ref{lemma:const-pred}.}, which is not the case for SAGE. While our approach aims to provide nuanced insights into variable importance relative to a specific set, SAGE aims to quantify the overall importance of variables for the model.\\
Feature importance relative to variables that have not been included in the training set has not been studied before. The indirect influence of variables that the model does not computationally rely but statistically depend on has been studied e.g. in \cite{Adler2018}.

\section{Background and Notation}

\subsection{Notation}

\input{figures/notation-overview}
We denote the target variable, i.e., the variable the model predicts, as $Y$ and feature variables by $X_{(.)}$. We refer to the variables as features to emphasize when they were used in model training. Their observations are denoted by $y$ and $x_{(.)}$. We use $D := \pset$ for the index set of all features included in model training and $j$ for the index of our feature of interest, $X_j$. The index set of the remaining variables is denoted as $R := D \backslash \{j\}$ (rest, remainder). The index set of features, relative to which the importance of $X_j$ is considered, is denoted as $G$. As $G$ can refer to any index set of variables, we denote its intersection with $R$ as $\Go = R \cap G$ and its complement as $\Ru = R \backslash G$. We denote the index set of conditioning variables that were not made available to the model during training as $\Gs = G \backslash R$.\\
In case we add new elements to the conditioning set $G$, we will denote this set as $N$. The set may include variables within and outside $D$. The respective components are denoted as  $\Ns = N \backslash R$ and as $\No = R \cap N$. The remainder of $R$ without $G$ and $N$ is denoted as  $\Ruu = \Ru \backslash N$. 
We denote perturbed variables of interest relative to $G$ as $\tXjg$. We refer to the original and perturbed probability distribution of $X_j$ as the observational and interventional distribution $P(X_j, \dots)$ and $P(\tXjg, \dots)$.
The inspected model is denoted as $f$, its prediction as $\hat{Y}$.
Independence of $Y$ and $X$ conditional on $Z$ is denoted using $X \idp Y | Z$, the respective conditional dependence as $X \dpd Y | Z$.

\subsection{Feature Importance}
\label{sec:background:fi}

Performance-based feature importance methods assess the relevance of a feature of interest $X_j$ by assessing the impact of a perturbation of $X_j$ on the model's performance. Local feature importance methods  focus on the importance of features for specific data points, whereas global feature importance methods assess the impact over the whole domain. In the following, we focus on global methods.\\
Global feature importance is computed according to the following general schemata:
$$\fimp_j = \risktj - \risk \text{ or } \fimp_j = \frac{\risktj}{\risk}$$
where we denote the original risk of the model and the risk after perturbing $X_j$ as $\risk$ and $\risktj$, respectively. For estimation, the true risk $\risk$ is replaced with the empirical risk $\riske$.
\\
Feature importance methods furthermore differ in how they perturb and whether they rely on retraining the model. While some methods retrain the model after the perturbation (e.g. LOCO, \cite{lei2018distribution}), others evaluate the impact of the perturbation on the same original model (e.g. \cite{Breiman2001rf, Strobl2008}). In this work, we focus on methods that avoid retraining.\\
For methods that avoid retraining, we observe a dichotomy between two general perturbation approaches: resampling that preserves the \textit{marginal} and resampling that preserves the \textit{conditional} distribution. Marginal resampling was originally proposed to compute perturbed versions of $X_j$ by permuting the observations $x_j^{(i)}$ within the sample \cite{Breiman2001rf}. The respective sample breaks the dependence between $X_j$ and $(Y, X_R)$ while preserving the marginal distribution $P(X_j)$. More recently, Model Reliance was proposed \cite{Fisher2019}, which takes the expectation over all possible permutations. Resampling from the marginal distribution has been criticized to introduce bias, in particular because it overestimates the importance of correlated variables \cite{Strobl2008}, resulting in incorrect feature rankings \cite{Tolosi2011}.
It also leads to extrapolation under dependent features  \cite{Hooker2019,Molnar2020}, i.e. conclusions about the model are being drawn using unrealistic data points on which the model was not trained. CFI, on the other hand, samples from the conditional distribution $P(X_j|X_R)$ \cite{Strobl2008,Tuv2009,Barber2015,Candes2018,Fisher2019,Hooker2019,Molnar2020}. A large variety of model-specific methods exist \cite{Gromping2009,Wei2015}. Conditional variants quantify the importance of a feature given the information that all remaining features $R$ contain about $X_j$ \cite{Molnar2020pitfalls}, thereby avoiding evaluation of the model on unrealistic datapoints \cite{Molnar2020}.\\

\section{Relative Feature Importance}
\label{sec:rfi}

Relative Feature Importance is a general framework that assesses feature importance relative to arbitrary variable sets $G$. The frameworks subsumes PFI and CFI as two extreme special cases.\\
In PFI, $X_j$ is replaced with a perturbed version that preserves the marginal distribution $P(X_j)$ while breaking the dependencies with $Y$ and all features. In CFI, a perturbed version of $X_j$ is used that preserves the conditional distribution $P(X_j|X_R)$, thereby only breaking conditional dependence between $X_j$ and $Y$ given all features. As our analysis in Section \ref{sec:interpreting-rfi} establishes, the replacement strategies of PFI and CFI define extreme reference points. CFI quantifies the contribution relative to \textit{all} remaining features $R$, whereas PFI regards a feature in isolation.\\
We go beyond the PFI versus CFI dichotomy. We argue that it is (1) meaningful (Section \ref{sec:interpreting-rfi}) and (2) practically useful (Section \ref{sec:examples}) to  replace $X_j$ with perturbed versions that preserve the conditional distribution $P(X_j|X_G)$ with respect to \textit{arbitrary} sets $G$ while requiring $\tXjg \Perp (\XRu, Y) | X_G$. $G$ can be a subset of $R$, but can also include variables not available at training time such that $G \backslash R \neq \emptyset$. We term the resulting method Relative Feature Importance (RFI):\\
\begin{definition}[Relative Feature Importance -- RFI] We define Relative Feature Importance with respect to a feature set $G$ with $Y \notin G$ and a fixed model $f$ as 
	$$\rfijg := \risktjg - \risk,$$
	where $\risktjg := \risk(Y, f(X_R, \tXjg))$ is the risk w.r.t.~to a replacement variable $\tXjg$ and $\risk = \risk(Y, f(X_j, X_R))$ refers to the original risk. The replacement variable has to satisfy
	\begin{itemize}
		\item $\tXjg \sim P(X_j|X_G)$ and
		\item $\tXjg \idp (\XRu, Y) | X_G$.\\
	\end{itemize}
	\label{def:rfi}
\end{definition}
In the following section, we discuss the semantic meaning of RFI. The estimation of RFI is discussed in Section~\ref{sec:estimation}.

\section{Interpreting Relative Feature Importance}
\label{sec:interpreting-rfi}

IML techniques aim to provide insight into the model and, possibly, into the underlying data generating mechanism. However, IML techniques themselves are subject to interpretation. The characterization of an IML method by its mathematical definition is computationally precise, but has limited aid in guiding users to make conclusions about the underlying model and data. In this section we provide a (non-comprehensive) list of interpretation rules for RFI, that \textit{characterize the method by how it behaves in its context}. This context includes \textit{both the model and the underlying data generating mechanism}. More specifically, we link RFI to (conditional) independence in the underlying data set as well as to whether the model's prediction $\Yh$ is constant in the argument $x_j$ for a fixed $x_R$. While RFI can be used for quantification of feature importance, we focus our analysis on relevance as a binary property and characterize relative feature relevance ($\text{RFI} \neq 0$). We show that the implicit notion of relevance of RFI is defined by the choice of $G$. By modifying the conditioning set $G$ beyond the PFI versus CFI dichotomy, we are able to gain insight into more nuanced aspects of the model and the data generating mechanism. The main results are given in Theorem~\ref{theorem:nonzero-rfi-implications} and Theorem~\ref{theorem:rfi-diff}. 
Furthermore, we highlight limitations stemming from the choice of the loss function $L$ and the model fit for the interpretation, which are, in our humble opinion, underrepresented in the current discussion.\\
We structure our analysis by taking the user's perspective and asking "What can we infer from relative feature relevance?".\\ 


\subsection{Implications of Relative Feature Relevance}
\label{sec:interpreting-rfi:without-assumptions}

In the following, we analyze the implications of RFI without further assumptions about model and data. We thereby distinguish between two levels of explanation. Relative feature relevance provides insight, both into \textit{model} and \textit{data}.
\\
\begin{theorem}
	If $RFI_j^G \neq 0$ then
	\begin{itemize}
		\item $X_j \dpd (Y, \XRu) | X_G$ in the underlying distribution (data level)
		\item $\tilde{X}_j \dpd \Yh | X_R$ w.r.t. the interventional distribution $P(X_j| X_G)P(X_G, \XRu) > 0$ (model level)\\
	\end{itemize}
\label{theorem:nonzero-rfi-implications}
\end{theorem}
We prove Theorem~\ref{theorem:nonzero-rfi-implications} in two steps. First, we assess the implications of the respective independence for the underlying data set (Lemma~\ref{lemma:relative-under-indpc}). Then, we assess the implications of the respective independence for the model (Lemma~\ref{lemma:const-pred}). The contrapositions yield Theorem~\ref{theorem:nonzero-rfi-implications}.\\
\begin{lemma}
	If  $X_j \idp (Y, \XRu) | X_G$ for any G with $Y \notin G$ then $RFI_j^G = 0$.\\
	\label{lemma:relative-under-indpc}
\end{lemma}
We base the proof of Lemma~\ref{lemma:relative-under-indpc} on the insight that (because the model $f$ is fixed) an equivalence in distribution implies an equivalence in risk (Proposition \ref{prop:same-risk}). Therefore conditions under which the interventional distribution $P(\tXjg, X_R, Y)$ coincides with the original distribution $P(X_j, X_R, Y)$ are sufficient for $RFI=0$.\\
\begin{proposition} If observational and interventional distribution coincide, then risks with and without perturbation are equal:
$$P(Y, X_j, X_R) = P(Y, \tXjg, X_R) \Rightarrow \risk(f) = \risktjg(f)$$
\label{prop:same-risk}
\end{proposition}
\begin{IEEEproof}[Proof of Proposition \ref{prop:same-risk}] Given that $P(Y, X_j, X_R) = P(Y, \tilde{X}_j, X_R)$ we can write
\begin{align*}
  \risk(f) &= \mathbb{E}_{Y, X_j, X_R}[L(Y, f(X_j, X_R))]\\
   &= \mathbb{E}_{Y, \tilde{X}_j, X_R}[L(Y, f(\tilde{X}_j, X_R))] = \tilde{\risk}(f).
\end{align*}
\end{IEEEproof}
We show next that the conditional independence $X_j \Perp (\XRu, Y) | X_G$ is a sufficient condition for identity of both distributions.\\
\begin{IEEEproof}[Proof of Lemma~\ref{lemma:relative-under-indpc}]
It holds that
\begin{align*}
	P(Y, X_j, \XRu, X_G) &= &P(X_j|Y, \XRu, X_G) P(Y, \XRu, X_G)\\
	& \eqtext{$X_j \Perp (\XRu, Y) | X_G$}  &P(X_j|X_G) P(Y, \XRu, X_G)\\
	& \eqtext{(def)}  &P(\tXjg|X_G)P(Y, \XRu, X_G)\\
	& = &P(\tXjg, Y, \XRu, X_G).
\end{align*}
Using Proposition \ref{prop:same-risk} we can infer that $RFI_j^G = 0$.\\
\end{IEEEproof}
So far, we have assessed implications for the underlying data generating mechanism. Next, we assess implications for the inspected model $f$.\\
\begin{lemma}
If $\tXjg \idp \Yh | X_R$ w.r.t. the interventional distribution $P(\tXjg, X_G, \XRu)$ then $\text{RFI}_j^G = 0$ for any $G$.\\
\label{lemma:const-pred}
\end{lemma}
\begin{IEEEproof}[Proof of Lemma~\ref{lemma:const-pred}]
If the prediction for an observation $(x_1, \dots, x_p)$ is independent of the value $x_j '$ w.r.t. the interventional distribution, the prediction is unaffected when replacing $x_j$ with any value $x_j '$ with $P(x_j '|X_G=x_G)P(X_G=x_G, \XRu=x_{\Ru})> 0$. Consequently, any sample from $\tXjg$ yields the same prediction.\\
Furthermore values $x_j '$ with nonzero probability over the interventional distribution also have nonzero probability over the observational distribution. The interventional distribution can be rewritten as
\begin{align*}
    P(\tXjg, X_G, \XRu) &= P(\tXjg| X_G, \XRu) P(X_G, \XRu)\\
    &= P(\tXjg| X_G)P(X_G, \XRu)\\
    &= P(X_j| X_G)P(X_G, \XRu).\\
\end{align*}
Similarly, the observational distribution can be factorized into $P(X_j|X_G, \XRu)P(X_G, \XRu)$. As $P(X_j|X_G, \XRu) > 0 \Rightarrow P(X_j|X_G) > 0$ (which can be derived from, e.g., the law of total probability) it follows that $P(\tXjg, X_G, \XRu) > 0 \Rightarrow P(X_j, X_G, \XRu) > 0$.\\
Consequently the prediction $\yh$ for any value $x_j$ with positive probability $P(X_j=x_j|X_R=x_R)$ is identical given unchanged $x_R$.\\
As the conditional distributions of $X_j$ and $\tXjg$ overlap and the distribution of $X_R$ is unaffected, the prediction $\hat{Y}$ is identical with and without perturbation. Therefore $\risk = \risktjg$ and $RFI_j^G=0$.\\
\end{IEEEproof}
To summarize, we have shown that independence on the dataset and on the model level respectively imply $\rfijg = 0$ and can thereby prove Theorem~\ref{theorem:nonzero-rfi-implications}.\\
\begin{IEEEproof}[Proof of Theorem~\ref{theorem:nonzero-rfi-implications}]
The result follows from contraposition of Lemma~\ref{lemma:relative-under-indpc} and contraposition of Lemma~\ref{lemma:const-pred}.\\
\end{IEEEproof}
Theorem~\ref{theorem:nonzero-rfi-implications} shows that nonzero $\rfijg$ implies dependencies between sets of variables on the model level as well as on the data level. Which dependencies are relevant for $\rfijg$ can be controlled with the conditioning set $G$. Consequently, the conditioning set $G$ determines the method's implicit definition of relevance.
I.e., on the data level, if $X_j \idp (\XRu, Y) | X_G$ holds, $\rfijg$ is zero irrespective of any other dependencies that may hold, e.g. with $X_G$ (Lemma~\ref{lemma:relative-under-indpc}). Nonzero RFI, a difference in performance on interventional and observational distribution, can only be caused by dependencies that have been destroyed in the interventional distribution, the dependencies with and via $X_G$ are preserved by the replacement $\tXjg$ and can therefore not be responsible for $\rfijg \neq 0$. Similarly, on the model level, $\tXjg \idp \Yh | X_R$ over the interventional distribution $P(X_j|X_G)P(X_G, \XRu)$ yields zero RFI (Lemma~\ref{lemma:const-pred}). The behavior of the model outside the domain in which it is evaluated is irrelevant for $RFI_j^G$. What domain the model is evaluated over depends on the choice of $G$.\\
Because we can control RFI's implicit definition of relevance with $G$, RFI allows more nuanced insights into model and data than PFI or CFI alone. In Theorem~\ref{theorem:nonzero-rfi-implications}, we aim to make the implicit definition of relevance explicit. On the data level, nonzero RFI implies the dependence of $X_j$ with the tuple $(Y, \XRu)$ given $X_G$ ($X_j \dpd (Y, \XRu) | X_G$). In order to understand the aforementioned dependence, using the graphoid axioms contraction and weak union \cite{pearl1985graphoids}, the equivalent formulation below can be adduced:
$$(X_j \dpd Y |X_G) \vee (X_j \dpd \XRu |X_G, Y).$$
At least one of the two conditional dependencies has to hold for nonzero $\rfijg$. The first dependence can be rephrased as: $X_j$ is informative of $Y$, even if we already know $X_G$. It is more difficult to make sense of the second dependence. 
Under dependent features $(X_j \dpd \XRu |X_G, Y)$, the distribution of $X_j$ with $\XRu$ is not preserved under perturbation $\tXjg$. In the interventional distribution $P(\tXjg, \XRu)$ observations that are improbable or impossible w.r.t. the observational distribution $P(X_j, \XRu)$ can be possible and probable (and vice versa). Consequently, in the interventional distribution the feature distribution differs from the observation feature distribution. Even if $X_j \idp Y | X_G$ holds, the model may perform suboptimally due to this distribution shift and cause $\rfijg$ nonzero\footnote{Let e.g. $X_1, X_2$ be perfectly correlated and independent of $Y$. Then adding $X_1 - X_2$ does not alter its prediction performance, unless the dependence between the variables is broken. Also see \cite{Hooker2019} for a discussion in PFI.}. If the conditioning set is a superset of $R$ ($G \supseteq R$), such that set of remaining variables $\XRu$ is empty, it holds that $(X_j \idp \XRu |X_G, Y)$. Therefore nonzero RFI must be attributed to $(X_j \dpd Y |X_G)$ for $G \supseteq R$.\\
On the model level, nonzero RFI implies that the model's predictions are conditionally dependent on $\tXjg$ given the remaining features $R$ are fixed. E.g. for a linear model that has coefficient zero for all terms involving $X_j$, this dependence would not be fulfilled, and $\rfijg$ would be zero (Lemma~\ref{lemma:const-pred}). The model is evaluated over the interventional distribution $P(X_j| X_G)P(X_G, \XRu) > 0$, which varies depending on $G$. If $G$ contains a nearly perfect correlate of $X_j$, $X_j$ can be reconstructed well. In contrast, if $G = \emptyset$, for every possible $x_R$ the model is evaluated over the whole marginal distribution of $X_j$. Although choosing a smaller set $G \subset R$ leads to extrapolation under dependent features, it allows more insight into the model's mechanism. For interpretation purposes like safety, this is highly desirable.\\
In the preceding paragraphs we have highlighted the importance of the conditioning set $G$ for the method's implicit notion of relevance and illustrated the results from Theorem~\ref{theorem:nonzero-rfi-implications}. We have argued that the conditioning set controls which potential dependencies can be responsible for nonzero $\rfijg$. The insights lead to a further, interesting application of RFI. By assessing the difference $\deltarfi = \rfijg - \rfijgn$ when modifying the conditioning set $G$ by adding new elements $N$, we are able to assess the role of the dependencies with variables in $N$ relative to a baseline $G$. While for $\rfijg$ only dependencies of $X_j$ with and via $G$ are preserved, for $\rfijgn$ also dependencies with and via $N$ are maintained. If $\deltarfi$ is nonzero, this change has to be due to dependencies involving $N$, but not $G$. We substantiate this claim with Theorem~\ref{theorem:rfi-diff}. In order for $\deltarfi$ to be positive, the dependence $X_j \not \Perp X_N | X_G$ has to hold.\\
\begin{theorem}
If the difference $\Delta RFI_j^{G \to G \cup N} = RFI_j^G$ - $RFI_j^{G\cup N} \neq 0$, then  $X_j \not \Perp X_N | X_G$.\\ 
\label{theorem:rfi-diff}
\end{theorem}
\begin{IEEEproof}[Proof of Theorem~\ref{theorem:rfi-diff}]
Under independence $X_j \idp X_n | X_G$ it holds that 
\begin{align*}
	P(\tXjg, Y, \XRuu, X_G, X_N) &= P(\tXjg|Y, \XRuu, X_G, X_N)P(Y, \XRuu, X_G, X_N)\\
	&\eqtext{(def $\tXjg$)} P(X_j|X_G)P(Y, \XRuu, X_G, X_N)\\
	&\eqtext{$X_j \idp X_n | X_G$} P(X_j|X_G, X_N)P(Y, \XRuu, X_G, X_N)\\
	&\eqtext{(def $\tXjgn$)} P(\tXjgn|X_G, X_N)P(Y, \XRuu, X_G, X_N)\\
	&\eqtext{(def $\tXjgn$)} P(\tXjgn|Y, X_G, X_N, X_{\Ruu})P(Y, \XRuu, X_G, X_N)\\
	&= P(\tXjgn, Y, \XRuu, X_G, X_N)\\
\end{align*}
The equality $P(\tXjg, Y, \XRuu, X_G, X_N) = P(\tXjgn, Y, \XRuu, X_G, X_N)$ implies $P(\tXjg, Y, X_R) = (\tXjgn, Y, X_R)$. Invoking Proposition~\ref{prop:same-risk} it holds that the corresponding risks $\risk^{j|G}$ and $\risk^{j|G \cup N}$ are equal. As $\text{RFI}_j^G - \text{RFI}_j^{G \cup N} = \risk^{j|G} - \risk^{j|G \cup N}$ it holds that $X_j \not \Perp X_n | X_G \Rightarrow \Delta RFI_j^{G \to G \cup N} = 0$.  Contraposition proves Theorem~\ref{theorem:rfi-diff}.\\
\end{IEEEproof}
 While nonzero $\rfijg$ as well as nonzero $\deltarfi$ have clear implications, interpreting zero $\rfijg$ or zero $\deltarfi$ is difficult. For example, we may be tempted to interpret $\rfijg = 0$ as conditional independence in the data. However, the general principle that absence of evidence is no evidence for absence also applies in the context of RFI. A dependence in the data may not be captured by the model when it has a poor fit and does not rely on the respective variable. Similarly, although $f$ may be optimal, a dependence in higher moments may simply not be modeled by $f$ or captured by the loss $L$. As all aforementioned causes of nonzero RFI are potentially sufficient, but not necessary, it is unclear which of the causes nonzero RFI can be attributed to. Furthermore, the related problem of conditional independence testing is provably hard \cite{Shah2018}.\\
The theoretical insights that we derive in this Section (Theorem~\ref{theorem:nonzero-rfi-implications} and \ref{theorem:rfi-diff}) are applied and illustrated in a simulation study in Section \ref{sec:examples}.\\

\section{Estimation and Testing}
\label{sec:estimation}

Estimating and sampling from the conditional distribution is in general difficult, especially in high-dimensional continuous settings. Various approaches for replacing $X_j$ with samples from its conditional distribution exist, e.g., knockoff approaches \cite{Barber2015,Candes2018,Romano2019}, imputation and weighting \cite{Fisher2019} or permutation within decision tree leaves \cite{molnar2020model}. We used Model-X knockoffs \cite{Candes2018} in this work, but note that the RFI approach is agnostic to its algorithmic implementation.\\
Using (standard) empirical risk estimates, our RFI estimate is
\begin{equation*}
  \hat{\text{RFI}}_j^G = \frac{1}{n}\sum_{i=1}^n L\left(\yi, f(\tilde{x}^{(i)}_j, x^{(i)}_R)\right) -  \frac{1}{n}\sum_{i=1}^n L\left(\yi, f(x^{(i)}_j, x^{(i)}_R)\right)
\end{equation*}
where $\tilde{x}^{(i)}_j$ is a sample from $\tXjg$. We can then test for nonzero $\text{RFI}_j^G$ using procedures for conditional independence tests, e.g., \cite{Watson2019}, thereby quantifying the uncertainty coming from empirical risk minimization.
Because of the central limit theorem, the empirical risk converges (in probability) to a Gaussian distribution with increasing number of observations. Therefore, one-sided, paired t-tests can be used to infer tests and confidence intervals \cite{Watson2019}.
The test procedures proposed in \cite{Watson2019} are agnostic to the conditioning set for the perturbation $\tXjg$.
For smaller samples, the Exact Test by Fisher may be used.\\
The t-test and Fisher Exact Test ignore uncertainty and bias of the estimation procedures, i.e. the ML model and the knockoff-sampler are treated as ``fixed". E.g. misspecified, suboptimal models may not capture dependencies. Or dependencies are in higher moments that are not captured by the loss. Consequently, without further assumptions, the framework does not provide a test for conditional independence in the dataset.\\
The popular testing procedures for knockoffs proposed by \cite{Candes2018} provide FDR over all features, but does not test the significance of the importance of individual features.\\

\section{Simulation Studies}
\label{sec:examples}

In the following, we demonstrate the usefulness of RFI on two simulation studies. In the first example, we use RFI to expose indirect influence of variables that are not computationally used by the model. In the second example, we assess feature importance relative to a confounder that was unavailable at training time. In both examples, we represent the underlying data generating mechanism, that gives rise to the dependencies in the data, with a causal directed acyclic graph (DAG). The code for the examples is available online\footnote{Link to Code: \href{https://github.com/gcskoenig/icpr2020-rfi}{\texttt{https://github.com/gcskoenig/icpr2020-rfi}}}.

\subsection{Indirect Influence}
\label{sec:examples:indirect-influence}

A prominent application of interpretable machine learning is auditing models regarding its reliance on protected attributes $A$ like age or sex. A reliance on the respective attributes may result in unfair discrimination and requires further inspection. With approaches like fairness through unawareness \cite{fairnessinmlbook}, the model does not rely on protected attributes directly. However, by implicitly reconstructing the sensitive attributes using seemingly harmless correlates, the model can indirectly make use of the protected attribute resulting in potentially harmful, unfair discrimination \cite{fairnessinmlbook}.\\
PFI and CFI cannot expose such indirect influence. As Lemma~\ref{lemma:const-pred} proves, $RFI_A^G$ is zero for a model that does not (directly) use the feature of interest $A$ for the prediction for any conditioning set $G$. Furthermore, from PFI and CFI alone, we cannot infer whether the importance of a variable can be attributed to its dependence with an indirect influence. Using $\rfijg$ with $G = A$ we preserve the influence of $A$ on the prediction and can thereby restrict the attribution of importance to contributions stemming from dependencies not involving $A$ (Theorem~\ref{theorem:nonzero-rfi-implications}, Lemma~\ref{lemma:relative-under-indpc}). The difference to $\deltarfi$ with $G  = \emptyset$ and $N = A$ exposes the indirect influence.\\
Not every indirect influence from a sensitive attribute is considered undesirable. Certain correlates of $A$ may indeed be valid criteria for a decision (e.g. \cite{Bonham2016}). Importance stemming from dependencies with $A$ via such resolving variables $Z$ would be considered acceptable. We can assess the indirect influence beyond contributions stemming from dependence via $Z$ by comparing to a baseline $G = Z$. In this baseline, contributions via $Z$ are preserved and therefore irrelevant for RFI. Consequently, when setting $N = A$, the difference $\deltarfi$ only quantifies indirect influence that is not resolved by $Z$.\\
We demonstrate the usefulness of RFI to expose indirect influence in a simulation study. The dataset is a sample drawn from the distribution induced by a structural causal model (SCM) depicted in Figure~\ref{fig:graph-example2}. All relationships are additive linear with coefficients $1$ and Gaussian noise terms ($\sigma_1 = \sigma_2 = \sigma_4 = 1$, $\sigma_3=0.3$ and $\sigma_y=0.5$). An ordinary least squares linear regression model was fit to predict $Y$ from $X_1, \dots, X4$ (MSE = $0.25$, $f(x_1, x_2, x_3, x_4) = 0.00x_1-0.01x_2+1.01x_3+1.00x_4$). We trained model-X knockoffs \cite{Candes2018} on the training data and evaluated RFI on test data. Sample size is $10^5$ with $10\%$ test data.\\
In order to quantify the direct influence of the features we compute PFI. As we can see in Figure~\ref{fig:rfi-chain}, $X_1$ and $X_2$ are considered irrelevant. In order to expose their indirect influence, we additionally compute RFI with respect to $G=\{X_1\}$ and $G=\{X_2\}$ respectively. For both variables we observe a drop in importance of $X_3$ and $X_4$. Consequently both $X_1$ and $X_2$ have an indirect influence on the target (Theorem~\ref{theorem:rfi-diff}).\\
Furthermore we are interested in whether the indirect influence of $X_1$ can be resolved by $X_2$. We therefore compute $\rfijgn$ with $G = \{X_2\}$ and $N = \{X_1\}$. We see that for $X_3$ no change in importance can be observed.  This is due to the independence $X_1 \idp X_3 | X_2$\footnote{As faithfulness and causal markov condition hold, $d$-separation in the graph and (conditional) independence coincide\cite{Pearl2009}. We can therefore read the independence structures off Figures~\ref{fig:graph-example2} and ~\ref{fig:graph-example1}.} (Theorem~\ref{theorem:rfi-diff}). The indirect influence is resolved. However, for $X_4$ the importance decreases further and is therefore not resolved by $X_2$. This is in alignment with the dependence $X_1 \dpd X_4 | X_2$ implied by the graph (Figure~\ref{fig:graph-example2}).\\
\input{figures/example2}
\input{figures/chain_figure}
\subsection{Variables Outside Training Set}
\label{sec:examples:confounder}
When designing a model $f$, a practitioner may have decided to exclude a variable from the feature set, e.g., because it was then considered irrelevant, it belongs to a different modality or would have required further preprocessing. Furthermore, when auditing a machine learning model $f$, variables that have not been available for the training of the model may be accessible.\\
In this example, we demonstrate that variables outside the training set can be included in the conditioning set for RFI. Consequently, importance of the features relative to variables outside the training set and the indirect influence of such variables can be assessed. More specifically, we simulate a hypothetical situation where the influence of a previously unknown confounder $C$ shall be evaluated. This variable $C$ is available for the model audit. In particular, we wonder whether the features $X_1$, $X_2$ and $X_3$ are only or partly important due to a dependence via $C$.\\
The dataset was sampled from a structural causal model (SCM) depicted in Figure~\ref{fig:graph-example1}. Assuming faithfulness and the causal Markov condition, this DAG implies the following (conditional) (in-)dependencies: $X_1$ is independent of $C$, $X_3$ is independent of $Y$ conditional on $C$, and $X_2$ is dependent on $Y$. Note that the dependence between $X_2$ and $Y$ is due to the common cause $C$ as well as due to a direct effect of $X_2$ on $Y$. All relationships are additive linear with coefficients $1$ and additive Gaussian  noise ($\sigma_1 = \sigma_2 = \sigma_C = 1.0$ and $\sigma_3 = \sigma_Y = 0.5$). We fit an ordinary least squares linear regression model on $X_1$, $X_2$ and $X_3$ to predict $Y$ (MSE = $0.40$, $f(x_1, x_2, x_3) = 1.0x_1+1.17x_2+0.67x_3$). $C$ was not available for model training. We trained Model-X knockoffs \cite{Candes2018} on training data and sampled from $\tXjg$ on test data. Sample size is $10^5$ with $10\%$ test data.\\
When computing $\text{RFI}_j^C$ ($G=\{C\}$) for each variable, the different relationships with $C$ become apparent. The respective results are depicted in Figure~\ref{fig:rfi-confounding}. For $X_1$ the feature importance relative to $C$ remains unchanged as the variables are pairwise independent (Theorem~\ref{theorem:rfi-diff}). For $X_3$, that is only dependent with $Y$ via $C$, it completely vanishes (Lemma~\ref{lemma:relative-under-indpc}). For $X_2$ the feature importance decreases but remains nonzero, as $X_2$ is dependent with $Y$ directly and via $C$.\\
Consequently, using RFI, we can (1) identify variables that are important due to a variable unavailable at training time and (2) distinguish between variables that only depend on $Y$ via $C$ from those that do not. With PFI ($G = \emptyset$) or CFI ($G = R$) such a distinction is in general not possible.\\
\input{figures/example1}
\input{figures/confounding_figure}
\section{Discussion}
We proposed relative feature importance (RFI), a general conditional feature importance framework which allows to condition on arbitrary sets of other features, including features outside the training set. We underpin the method with theoretical results allowing insight into both model and underlying dataset. In a simulation study, the usefulness of the method for the exposure of indirect influence is demonstrated.\\
Relative feature importance requires sampling from (unknown) conditional distributions. For continuous variables and in high-dimensional settings this task is challenging and an open area of research \cite{Candes2018, Romano2019}. Uncertainty stemming from inaccurate sampling may affect the interpretation. The quality of insight into the underlying dataset strongly depends on the training and evaluation of the model. Dependencies in higher moments are usually not modeled and not captured by standard loss functions and can therefore not be detected. Especially the interpretation of zero RFI requires careful assessment of the model specification. Further research is needed to assess necessary assumptions for the interpretation of RFI. These challenges are not unique to RFI, but apply more generally in the field of interpretable machine learning \cite{Molnar2020pitfalls}.

\bibliography{root}
\bibliographystyle{plain}

\end{document}

%% file: latex-math/basic-ml.tex
\newcommand{\pset}{\{1, \ldots, p\}}                                        
\renewcommand{\xi}[1][i]{\mathbf{x}^{(#1)}}                                          
\newcommand{\yi}[1][i]{y^{(#1)}}                                            

\newcommand{\yh}{\hat{y}}                                                   

\newcommand{\risk}{\mathcal{R}}                                             
\newcommand{\riske}{\mathcal{R}_{\text{emp}}}                               



%% file: notation.tex
\newcommand{\fimp}{\text{FI}}
\newcommand{\rfi}{\text{RFI}}
\newcommand{\rfijg}{\rfi_j^G}
\newcommand{\rfijgn}{\rfi_j^{G \cup N}}
\newcommand{\deltarfi}{\Delta RFI_j^{G \to G \cup N}}

\newcommand{\risktj}{\tilde{\risk}^j}
\newcommand{\risktjg}{\tilde{\risk}^{j|G}}

\newcommand{\Ru}{\underline{R}}
\newcommand{\Ruu}{\underline{\underline{R}}}
\newcommand{\Go}{\overline{G}}
\newcommand{\No}{\overline{N}}
\newcommand{\Gs}{G^*}
\newcommand{\Ns}{N^*}

\newcommand{\tXjg}{\tilde{X}_j^G}
\newcommand{\tXjgn}{\tilde{X}_j^{G \cup N}}

\newcommand{\XRu}{X_{\Ru}}
\newcommand{\XRuu}{X_{\Ruu}}

\newcommand{\Yh}{\hat{Y}}

\newcommand{\idp}{\Perp}
\newcommand{\dpd}{\not \idp}

%% file: figures/notation-overview.tex
\begin{figure}[H]
	\centering
	\begin{tikzpicture}[thick, scale=.9, every node/.style={scale=0.8, line width=0.3mm, black, fill=white}]
		\draw[fill=green, opacity=0.3, draw opacity=0.0] (-1.5,0) circle (0.4cm);
		\draw[fill=green, opacity=0.3, draw opacity=0.0] (0,0) circle (1cm);
		\draw[fill=orange, opacity=0.3, draw opacity=0.0] (1,0) circle (0.75cm);

		\node[font=\large, fill opacity=0.0, text opacity=1.0] (xj) at  (-1.5,-0.65) {$\{j\}$};
		\node[font=\large, fill opacity=0.0, text opacity=1.0] (R) at  (0,-1.25) {$R$};
		\node[font=\large, fill opacity=0.0, text opacity=1.0] (uR) at  (-0.25,-.04) {$\underline{R}$};
		\node[font=\large, fill opacity=0.0, text opacity=1.0] (G) at  (1,-1) {$G$};
		\node[font=\large, fill opacity=0.0, text opacity=1.0] (Gstar) at  (1.35,0) {$G^*$};
		\node[font=\large, fill opacity=0.0, text opacity=1.0] (oG) at  (.7,.04){$\overline{G}$};

	\end{tikzpicture}
	\caption{Overview of our notation.}
\label{fig:scm-data-model} 
\end{figure}
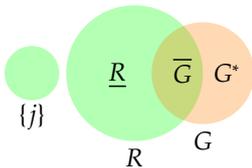

%% file: figures/example2.tex
\begin{figure}[h]
	\centering
	\begin{tikzpicture}[thick, scale=1.5, every node/.style={scale=0.9, line width=0.3mm, black, fill=white}]

		\node[draw, circle, scale=0.7] (x1) at (-2, 0) {$X_1$};
		\node[draw, circle, scale=0.7] (x2) at (-1, .3) {$X_2$};
		\node[draw, circle, scale=0.7] (x3) at (0, .3) {$X_3$};
		\node[draw, circle, scale=0.7] (x4) at (-0.5, -.3) {$X_4$};
		\node[draw, circle, scale=0.7] (y) at (1,0) {$Y$};
		
		\draw[->] (x1) -- (x2);
		\draw[->] (x2) -- (x3);
		\draw[->] (x1) -- (x4);
		\draw[->] (x3) -- (y);
		\draw[->] (x4) -- (y);

	\end{tikzpicture}
	\caption{Variable $X_1$ influences $Y$ both via the chain $X_2 \rightarrow X_3$ and via $X_4$. $X_1$ may be some undesired influence, and $X_2$ a variable resolving the undesired influence. We find that the prediction can nevertheless be influenced via $X_4$ by comparing $RFI_4^{X_2}$ with $RFI_4^{X_2, X_1}$ (Figure \ref{fig:rfi-chain}). All relationships are additive linear Gaussian with all coefficients being equal to $1$ and $\sigma_1 = \sigma_2 = \sigma_4 = 1$, $\sigma_3=0.3$ and $\sigma_y=0.5$.}
\label{fig:graph-example2} 
\end{figure}
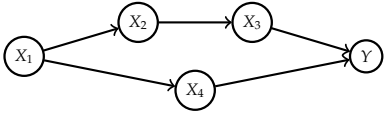

%% file: figures/chain_figure.tex
\begin{figure}[H]
    \centering
    \includegraphics[width=1.0\linewidth]{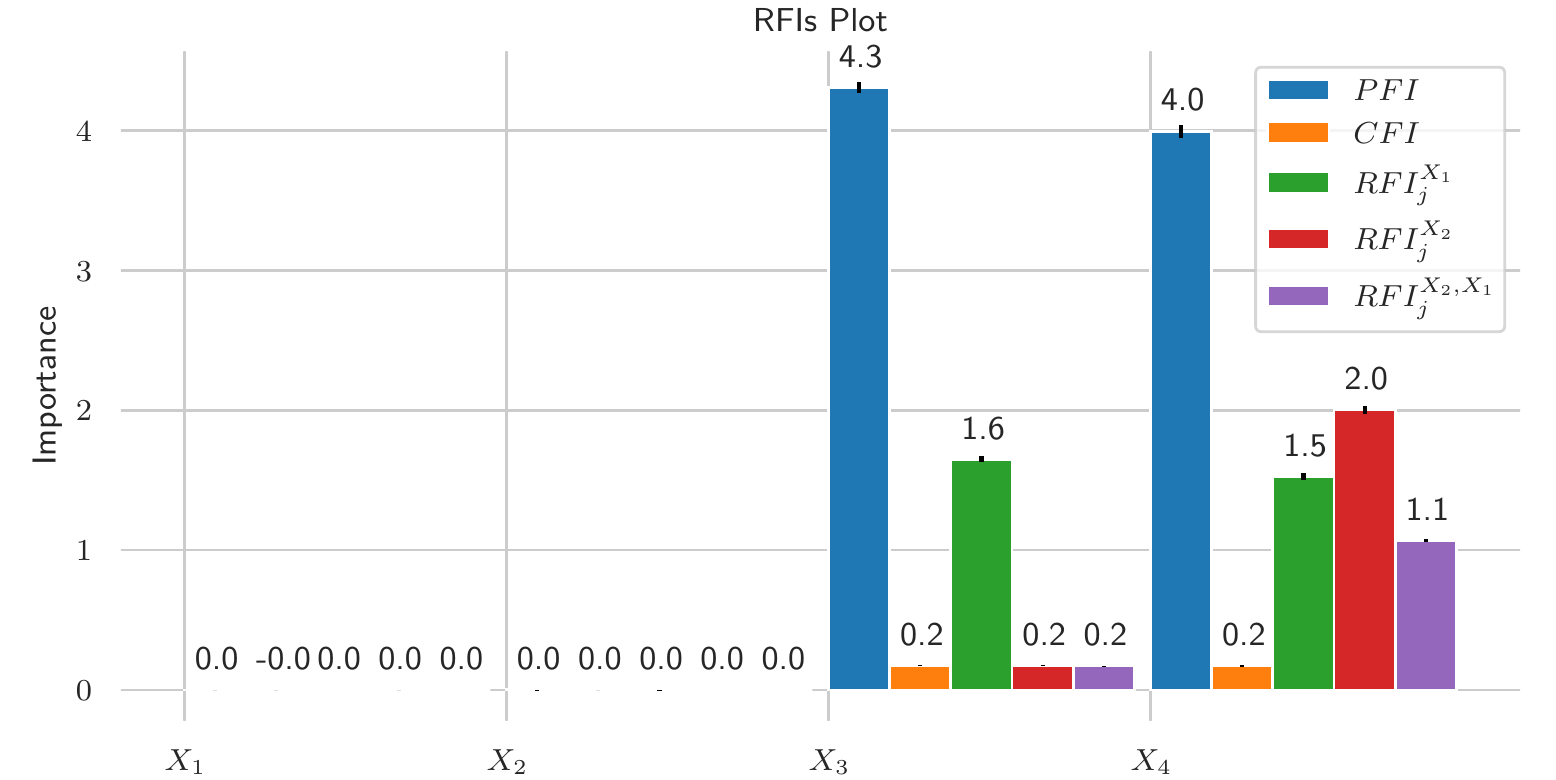}
    \caption{RFI's for a linear regression model fitted on the dataset illustrated in Figure \ref{fig:graph-example2}. Feature importance values are averaged over 30 runs and rounded. Feature importance values are averaged over 30 runs and rounded. We evaluated significance using a t-test for the first run. All positive features were significant at $\alpha=0.01$, whereas for all zero RFI values the null could not be rejected.
    For $X_1$ and $X_2$ all RFIs are zero, whereas for $X_3$ and $X_4$ RFIs are positive. We see that $X_1$ and $X_2$ both have an indirect influence on $X_3$ and $X_4$, but that $X_2$ can resolve the influence of $X_1$ on $X_3$.}
    \label{fig:rfi-chain}
\end{figure}

%% file: figures/example1.tex
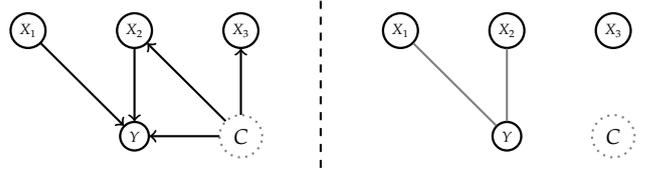
\begin{figure}[h]
	\centering
	\begin{tikzpicture}[thick, scale=1.4, every node/.style={scale=0.8, line width=0.3mm, black, fill=white}]

		\node[circle] (c) at (0, 0) {$C$};
		\draw[gray, dotted] (c) circle (0.2	);
		\node[draw, circle, scale=0.7] (x1) at (-2, 1.) {$X_1$};
		\node[draw, circle, scale=0.7] (x2) at (-1, 1.) {$X_2$};
		\node[draw, circle, scale=0.7] (x3) at (0, 1.) {$X_3$};
		\node[draw, circle, scale=0.7] (y) at (-1, 0) {$Y$};
		
		\draw[->] (c) -- (y);
		\draw[->] (c) -- (x2);
		\draw[->] (c) -- (x3);
		
		\draw[->] (x1) -- (y);
		\draw[->] (x2) -- (y);

		\node[circle] (c) at (3.5, 0) {$C$};
		\draw[gray, dotted] (c) circle (0.2	);
		\node[draw, circle, scale=0.7] (x1) at (1.5, 1.) {$X_1$};
		\node[draw, circle, scale=0.7] (x2) at (2.5, 1.) {$X_2$};
		\node[draw, circle, scale=0.7] (x3) at (3.5, 1.) {$X_3$};
		\node[draw, circle, scale=0.7] (y) at (2.5, 0) {$Y$};
				
		\draw[-, gray] (x1) -- (y);
		\draw[-, gray] (x2) -- (y);
		
		\draw[dashed](0.75, -.3) -- (0.75, 1.3);
	\end{tikzpicture}
	\caption{\textit{Left:} We see the causal graph $\mathcal{G}$ corresponding to the Structural Causal Model that was used to generate the dataset used in Figure \ref{fig:rfi-confounding}. All relationships are additive linear Gaussian with all coefficients equal to $1$ and $\sigma_1 = \sigma_2 = \sigma_C = 1.0$ and $\sigma_3 = \sigma_Y = 0.5$. \textit{Right:} Pairwise dependencies after conditioning on $C$.}
\label{fig:graph-example1} 
\end{figure}

%% file: figures/confounding_figure.tex
\begin{figure}[H]
    \centering
    \includegraphics[width=0.7\linewidth]{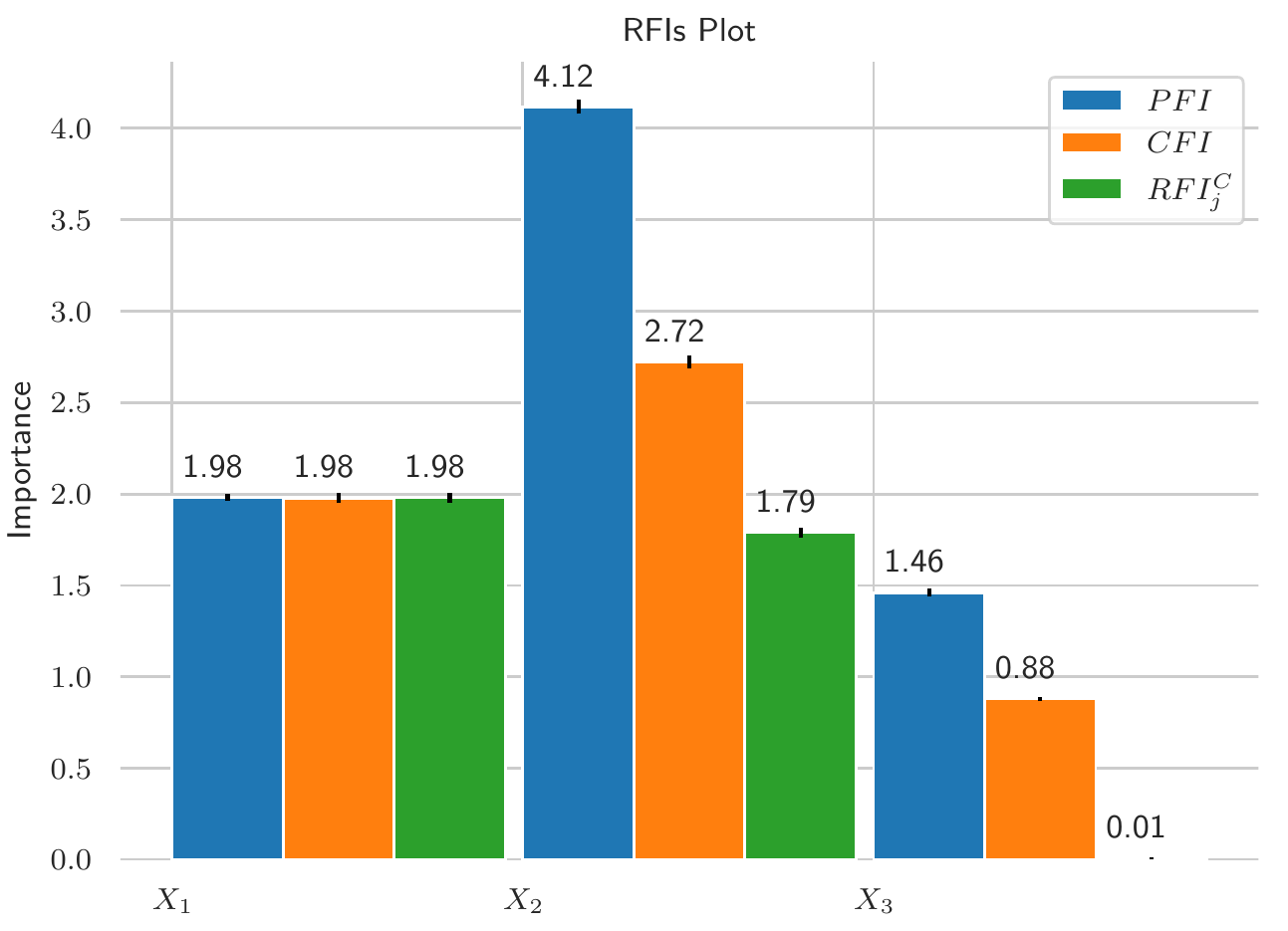}
    \caption{Feature Importance results corresponding to the dataset depicted in Figure \ref{fig:graph-example1}. We averaged RFI over 30 runs. RFI for $X_1$ is unaffected by changes in $G$, for $X_2$ RFI drops with $C$ is added to $G$. For $X_3$ RFI vanishes relative to $C$. For all except for $RFI_{X_3}^C$ the null can be rejected at $\alpha=0.01$ in the first run.}
    \label{fig:rfi-confounding}
\end{figure}

%% file: root.bbl
\begin{thebibliography}{10}

\bibitem{Adler2018}
Philip Adler, Casey Falk, Sorelle~A. Friedler, Tionney Nix, Gabriel Rybeck,
  Carlos Scheidegger, Brandon Smith, and Suresh Venkatasubramanian.
\newblock Auditing black-box models for indirect influence.
\newblock {\em Knowledge and Information Systems}, 54(1):95--122, 2018.
\newblock arXiv: 1602.07043.

\bibitem{Barber2015}
Rina~Foygel Barber, Emmanuel~J Candès, and {others}.
\newblock Controlling the false discovery rate via knockoffs.
\newblock {\em The Annals of Statistics}, 43(5):2055--2085, 2015.
\newblock Publisher: Institute of Mathematical Statistics.

\bibitem{fairnessinmlbook}
Solon Barocas, Moritz Hardt, and Arvind Narayanan.
\newblock {\em Fairness and Machine Learning}.
\newblock fairmlbook.org, 2019.
\newblock \url{http://www.fairmlbook.org}.

\bibitem{Bonham2016}
Vence~L Bonham, Shawneequa~L Callier, and Charmaine~D Royal.
\newblock Will precision medicine move us beyond race?
\newblock {\em The New England journal of medicine}, 374(21):2003, 2016.

\bibitem{Breiman2001rf}
Leo Breiman.
\newblock Random forests.
\newblock {\em Machine Learning}, pages 1--122, 2001.

\bibitem{Budescu1993}
David~V Budescu.
\newblock Dominance analysis: a new approach to the problem of relative
  importance of predictors in multiple regression.
\newblock {\em Psychological bulletin}, 114(3):542, 1993.

\bibitem{Candes2018}
Emmanuel Candès, Yingying Fan, Lucas Janson, and Jinchi Lv.
\newblock Panning for gold: ‘model-{X}’ knockoffs for high dimensional
  controlled variable selection.
\newblock {\em Journal of the Royal Statistical Society. Series B: Statistical
  Methodology}, 80(3):551--577, 2018.
\newblock arXiv: 1610.02351.

\bibitem{Covert2020}
Ian Covert, Scott Lundberg, and Su-In Lee.
\newblock Understanding {Global} {Feature} {Contributions} {Through} {Additive}
  {Importance} {Measures}.
\newblock {\em arXiv preprint arXiv:2004.00668}, 2020.

\bibitem{Dastin2018}
Jeffrey~(Reuters) Dastin.
\newblock Amazon scraps secret {AI} recruiting tool that showed bias against
  women.
\newblock {\em Reuters}, 2018.

\bibitem{Torre2018}
Lydia de~la Torre.
\newblock A {Guide} to the {California} {Consumer} {Privacy} {Act} of 2018.
\newblock {\em SSRN Electronic Journal}, pages 1--17, 2018.

\bibitem{Dressel2018}
Julia Dressel and Hany Farid.
\newblock The accuracy, fairness, and limits of predicting recidivism.
\newblock {\em Science Advances}, 4(1):1--6, 2018.

\bibitem{Fisher2019}
Aaron Fisher, Cynthia Rudin, and Francesca Dominici.
\newblock All models are wrong, but many are useful: Learning a variable's
  importance by studying an entire class of prediction models simultaneously.
\newblock {\em Journal of Machine Learning Research}, 20(177):1--81, 2019.

\bibitem{Gromping2009}
Ulrike Grömping.
\newblock Variable importance assessment in regression: linear regression
  versus random forest.
\newblock {\em The American Statistician}, 63(4):308--319, 2009.
\newblock Publisher: Taylor \& Francis.

\bibitem{Hooker2019}
Giles Hooker and Lucas Mentch.
\newblock Please {Stop} {Permuting} {Features}: {An} {Explanation} and
  {Alternatives}.
\newblock {\em arXiv preprint arXiv:1905.03151v}, pages 1--15, 2019.
\newblock arXiv: 1905.03151v1.

\bibitem{lei2018distribution}
Jing Lei, Max G’Sell, Alessandro Rinaldo, Ryan~J Tibshirani, and Larry
  Wasserman.
\newblock Distribution-free predictive inference for regression.
\newblock {\em Journal of the American Statistical Association},
  113(523):1094--1111, 2018.

\bibitem{Lipovetsky2001}
Stan Lipovetsky and Michael Conklin.
\newblock Analysis of regression in game theory approach.
\newblock {\em Applied Stochastic Models in Business and Industry},
  17(4):319--330, 2001.
\newblock Lipovetsky2001.

\bibitem{Lundberg2017}
Scott~M. Lundberg and Su~In Lee.
\newblock A unified approach to interpreting model predictions.
\newblock {\em Advances in Neural Information Processing Systems},
  2017-Decem(Section 2):4766--4775, 2017.
\newblock arXiv: 1705.07874.

\bibitem{molnar2020model}
Christoph Molnar, Gunnar K{\"o}nig, Bernd Bischl, and Giuseppe Casalicchio.
\newblock Model-agnostic feature importance and effects with dependent
  features--a conditional subgroup approach.
\newblock {\em arXiv preprint arXiv:2006.04628}, 2020.

\bibitem{Molnar2020}
Christoph Molnar, Gunnar K{\"o}nig, Bernd Bischl, and Giuseppe Casalicchio.
\newblock Model-agnostic feature importance and effects with dependent
  features-a conditional subgroup approach.
\newblock {\em arXiv preprint arXiv:2006.04628}, 2020.

\bibitem{Molnar2020pitfalls}
Christoph Molnar, Gunnar K{\"o}nig, Julia Herbinger, Timo Freiesleben, Susanne
  Dandl, Christian~A. Scholbeck, Giuseppe Casalicchio, Moritz Grosse-Wentrup,
  and Bernd Bischl.
\newblock Pitfalls to avoid when interpreting machine learning models.
\newblock {\em arXiv preprint arXiv:2007.04131}, 2020.

\bibitem{Pearl2009}
Judea Pearl.
\newblock {\em Causality}.
\newblock Cambridge university press, 2009.

\bibitem{pearl1985graphoids}
Judea Pearl and Azaria Paz.
\newblock {\em Graphoids: A graph-based logic for reasoning about relevance
  relations}.
\newblock University of California (Los Angeles). Computer Science Department,
  1985.

\bibitem{Romano2019}
Yaniv Romano, Matteo Sesia, and Emmanuel Candès.
\newblock Deep knockoffs.
\newblock {\em Journal of the American Statistical Association}, pages 1--12,
  2019.
\newblock Publisher: Taylor \& Francis.

\bibitem{Shah2018}
Rajen~D Shah and Jonas Peters.
\newblock The hardness of conditional independence testing and the generalised
  covariance measure.
\newblock {\em arXiv preprint arXiv:1804.07203}, 2018.

\bibitem{Strobl2008}
Carolin Strobl, Anne~Laure Boulesteix, Thomas Kneib, Thomas Augustin, and Achim
  Zeileis.
\newblock Conditional variable importance for random forests.
\newblock {\em BMC Bioinformatics}, 9:1--11, 2008.

\bibitem{Tolosi2011}
Laura Toloşi and Thomas Lengauer.
\newblock Classification with correlated features: unreliability of feature
  ranking and solutions.
\newblock {\em Bioinformatics}, 27(14):1986--1994, 2011.
\newblock Publisher: Oxford University Press.

\bibitem{Topol2019}
Eric~J Topol.
\newblock High performance medicine: the convergence of human and artificial
  intelligence.
\newblock {\em Nature Medicine}, 25(January), 2019.
\newblock Publisher: Springer US.

\bibitem{Tuv2009}
Eugene Tuv, Alexander Borisov, George Runger, and Kari Torkkola.
\newblock Feature selection with ensembles, artificial variables, and
  redundancy elimination.
\newblock {\em Journal of Machine Learning Research}, 10(Jul):1341--1366, 2009.

\bibitem{Voigt2017}
Paul Voigt and Axel dem Bussche.
\newblock The eu general data protection regulation (gdpr).
\newblock {\em A Practical Guide, 1st Ed., Cham: Springer International
  Publishing}, 2017.
\newblock Publisher: Springer.

\bibitem{Watson2019}
David~S. Watson and Marvin~N. Wright.
\newblock Testing {Conditional} {Independence} in {Supervised} {Learning}
  {Algorithms}.
\newblock {\em arXiv preprint arXiv:1901.09917}, 2019.
\newblock arXiv: 1901.09917.

\bibitem{Wei2015}
Pengfei Wei, Zhenzhou Lu, and Jingwen Song.
\newblock Variable importance analysis: a comprehensive review.
\newblock {\em Reliability Engineering \& System Safety}, 142:399--432, 2015.
\newblock Publisher: Elsevier.

\bibitem{Xia2017}
Yufei Xia, Chuanzhe Liu, Yu~Ying Li, and Nana Liu.
\newblock A boosted decision tree approach using {Bayesian} hyper-parameter
  optimization for credit scoring.
\newblock {\em Expert Systems with Applications}, 78:225--241, 2017.
\newblock Publisher: Elsevier Ltd.

\bibitem{vstrumbelj2014explaining}
Erik Štrumbelj and Igor Kononenko.
\newblock Explaining prediction models and individual predictions with feature
  contributions.
\newblock {\em Knowledge and information systems}, 41(3):647--665, 2014.
\newblock Publisher: Springer.

\end{thebibliography}
